# The Neural Correlates of Linguistic Structure Building: Comments on Kazanina & Tavano (2022)

Nai Ding

Zhejiang University

A recent perspective paper by Kazanina & Tavano (referred to as the KT perspective in the following) argues how neural oscillations cannot provide a potential neural correlate for syntactic structure building. The view that neural oscillations can provide a potential neural correlate for syntactic structure building is largely attributed to a study by Ding, Melloni, Zhang, Tian, and Poeppel in 2016 (referred to as the DMZTP study).

The KT perspective is thought provoking, but has severe misinterpretations about the arguments in DMZTP and other studies, and contains contradictory conclusions in different parts of the perspective, making it impossible to understand the position of the authors. In the following, I summarize a few misinterpretations and inconsistent arguments in the KT perspective, and put forward a few suggestions for future studies.

## 1. Misinterpretations about the DMZTP study
### 1.1. Neural activity can only track linear constituents?

The most basic idea in DMZTP is that the brain encodes hierarchically embedded structures. We can first consider a simple condition in which two words (e.g., X and Y) combine into a structure $[X\ Y]_{P1}$, which is further combined with a word Z to form a bigger structure $[\ [X\ Y]_{P1}\ Z]_{P2}$. Examples for such a structure include, e.g., "the boy laughed" and "my daughter smiled". The basic hypothesis in DMZTP is that there is neural activity tracking P1 as well as neural activity tracking P2. Here, tracking means that the waveform of recorded neural activity shows signatures commensurate with the duration of the linguistic structure. The DMZTP study only considered relatively simple structures similar to [ [X Y]$_{P1}$ Z]$_{P2}$, with P1 being referred to as phrases and P2 being referred to as sentences.

The KT perspective, however, somehow interprets the idea to be the opposite, i.e., "[syntactic constituents form a linear sequence in which a constituent begins where another constituent ends]". This confusion can potentially be alleviated if the "phrases" and "sentences" were referred to "structure level 1" and "structure level 2" in the DMZTP study (if the terminology does not lead to the confusion that only two levels of embedding is allowed). The hypothesis in DMZTP is that neural activity can track each linguistic structure that is mentally constructed.

## 1.2. Neural activity can only track of 500-ms long phrases?

The KT perspective seems to argue that the DMZTP study and other studies tend to believe that all phrases have similar duration around 500 ms. The reason to generate this rather counterintuitive hypothesis is that one experiment in DMZTP sets the phrase duration to 500 ms. The KT perspective ignores the fact that even within the same DMZTP study some other experiments present sentences with duration varying between 1 and 2 seconds (Fig. 4). Furthermore, a follow-up study explicitly manipulated the syllable rate so that the same phrases are presented at either 1 Hz or 2 Hz, to dissociate neural tracking of phrases from potential signatures of intrinsic neural activity at 1 or 2 Hz (Sheng et al., 2019). As expected, whether the phrases are presented at 1 or 2 Hz, cortical activity can track the phrases and the cortical networks generating the phrase-tracking activity remain similar. Furthermore, a recent study directly investigated the frequency range in which the neural tracking of linguistic structures can occur by temporally compressing speech, and the conclusion is neural activity can track sentences even when the sentence rate is as fast as 2-3 Hz (Lo, 2022). Therefore, in contrast to what the KT perspective suggests, there is no evidence that cortical activity can only track linguistic structures that last about 500 ms.

We believe the misinterpretation in the KT perspective is partly caused by the loose use of the terms such as "delta band" in the literature. In many studies mentioned by the KT perspective, the term "delta band" simply refers to slow neural activity that is typically below, e.g., 4 Hz. As far as we know, none of the studies reviewed in the KT perspective specify a potential neurophysiological mechanism to generate a "delta oscillation" and how the mechanism fail to track linguistic structures in, e.g., the theta and sub-delta band. Although the KT perspective seem to suggest fundamentally different neurophysiological mechanisms to track linguistic structures in the theta, delta, and sub-delta band, no single example mechanism is provided, rendering the argument unsubstantial.

## 2. Contradictory arguments
### 2.1. on the role of working memory
In the KT perspective, the most constructive suggestions for future research are in the last paragraph.

*"Although the claim that one should be mapping neural operations to parsing steps may meet with little disagreement, practical steps towards it are largely non-existent.* **As a notable exception**, *Nelson and colleagues demonstrated that…* **the high-gamma power … decreased whenever a snippet of syntactic structure was completed and cleared out of working memory.** *Hence, investigating how high-gamma activity is generated may be a step towards elucidating the mechanistic aspects of syntactic structure building by the brain. The time is ripe for the field to take concrete steps in this direction."*

This approach contrasts with the approach used by, e.g., DMZTP:

"Owing to the nature of the linguistic stimuli, **the spectral peaks at 2 Hz and 1 Hz may simply reflect an evoked response corresponding to the parser's regular building of phrases and sentences and clearing them out of working memory** (figure 4 in Ding et al. illustrates the response to phrases in the time domain and supports this interpretation). Hence, the rhythmic activity found in Ding et al.'s study and similar studies may be "a by-product of the existence of hierarchical units in language, not the mechanism by which those units come to exist"."

After reading both arguments, it seems like the main message is that both the correct and the incorrect approach to study the neurophysiological correlate of linguistic structures should focus on how completed structures are cleared out of working memory.

## 2.2. on the duration of phrases

A significant portion of the KT perspective is to correctly argue that phrases are not always 500 ms long, although nobody seems to have this belief when not speaking in the context of a particular experimental design. On the other hand, the KT perspective seems to believe that the phrases are always about 500 ms long in many of its argument, which is really confusing. For example, how could one or two phrases only have roughly a dozen words in the following argument?

*"On the working memory front, classic two-stage parsing models assume that, owing to working memory capacity limits, the parser shunts out parsed material corresponding to **one or two phrases roughly every half a dozen words as the outcome of the first stage** (as discussed by Fodor, prosodic factors may interact with the parser's decision to shunt a phrase out). **This shunting out thus happens rhythmically and falls within a delta rhythm.**"*

## 3. Suggestions

### 3.1 What kinds of neural activity should be focused on when searching for the neural correlate of linguistic structures

The question discussed in the KT perspective paper, i.e., whether neural oscillations contribute to syntactic structure building, is particularly challenging. On the one hand, connectionism research, including the current deep neural network approach, generally denies the existence of syntactic structure. On the other hand, whether the speech-tracking EEG/MEG response is related to neural oscillations is a heavily debated topic. In other words, although KT discusses the link between two concepts, i.e., neural oscillations and linguistic structure building, there are in fact intense debates about whether the two concepts themselves are meaningful or not.

These great challenges are of course not reasons to stop investigating the neural basis of sentence comprehension. Nevertheless, in my opinion, studies in this field should not tackle problems at all possible levels at the same time. According to the division of David Marr, cognitive neuroscientific studies can be divided into 3 levels, i.e., computational, algorithmic, and implementational levels. For example, whether the speech-tracking response is generated by evoked responses or phase resetting of ongoing neural oscillations is a question at the level of neural implementation. This question is an important neuroscience question and in principle can constrain the kind of algorithms can be allowed for syntactic analysis. Nevertheless, research on this question is currently at a primitive stage in which evoked responses and oscillations can hardly be distinguished and the neural mechanisms for either of them are largely unknown. Therefore, I suggest to not have a biased view on which kind of neural activity is relevant, e.g., evoked or oscillatory, when looking for the neural correlates of sentential processing. Instead, more efforts should be devoted to exploring the algorithmic level questions (in the sense of Marr), e.g., what kind of processing steps and representations can explain the spatial or temporal dynamics of the observed neural response to a sentence. For example, among the following questions, the field can benefit much more from investigations on questions 1 and 2 than debates on questions 3 and 4.

Question 1: Can some neural activity, e.g., low-frequency activity, high-gamma power, or spiking rate, track very long or very brief phrases?

Question 2: How does syntactic complexity, prosody, and semantics influence, e.g., neural tracking of phrases.

Question 3: Should the neural response tracking a 1-s long phrase be called delta-band or sub-delta-band activity?

Question 4: Should the neural response phase-locking to syllables/phrases be called neural activity entrained/synchronized to such units or neural activity tracking/encoding such units?

## 3.2 Space and time during the neural processing of sentences

MEG and EEG have superb time resolution but relatively low spatial resolution. Consequently, MEG and EEG studies often focus on the temporal dynamics of neural activity, which might have led to an illusion that there is a single neural oscillator or evoked-response generator inside the brain that oversees, e.g., speech processing. This is only an illusion since it begs the question of where the oscillator is and how this oscillator coordinates with the distributive semantic representations in the brain. MEG and EEG cannot resolve the neural populations encoding each linguistic unit but it does not mean that they do not exist. In fact, most models that attempt to explain the MEG/EEG neural tracking of linguistic structure assumes that different linguistic structures are encoded by different neural populations (e.g., the Ding 2020 model illustrated below, as well as the DORA and VS-BIND models discussed in the KT perspective). With these models in mind, a lot of confusions can be avoided, such as KT's five arguments challenging the "oscillations for chunking" view.

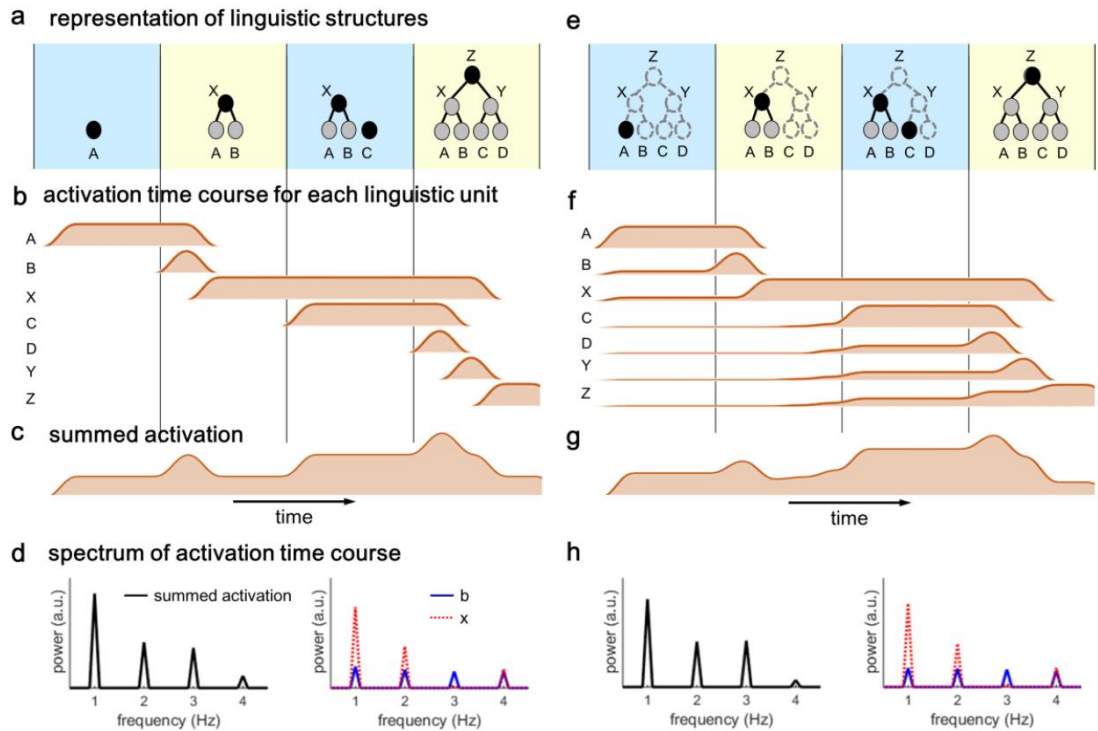

Figure 2 from Ding (2020). Evolution of the representations of different linguistic units in the processing buffer, for the sentence "fun games waste time". Panels a-d rely on a bottom-up parsing mechanism while panels e-h rely on a predicative top-down parsing mechanism. Panels a & e: The mental representation of the speech linguistic structure after hearing each word. The black nodes are maintained in the processing buffer while the gray nodes are stored in activated long-term memory. The hollow nodes with dashed lines indicate expected nodes, which are partially activated. The figures only show the final linguistic structure being derived after hearing each word, not showing the derivation process. Labels A-D represent the four words "fun", "games", "waste", "time", and labels X-Z represent phrases and sentences. Panels b & f: Time course of the neural representation of each linguistic unit in the processing buffer. Panels c & g: The sum of neural activation time courses in panels b & f. The summed time course can be viewed as the neural response recorded by a technique that cannot spatially resolve the neural representation of different linguistic units. Panels d & h: The power spectrum of the summed neural activation in panel b & f, and the power spectra of the activation of two representative node, i.e., node b & x. In this simulation, each word is set to 250 ms in duration and 10 sentences are sequentially played with no gap in between.